%% file: root.tex
\documentclass[letterpaper, 10 pt, conference]{ieeeconf}  % Comment this line out if you need a4paper
\usepackage{amsmath}
\usepackage{amssymb}
\usepackage{graphicx}
\usepackage{cleveref}
\usepackage{multirow} 
\usepackage{subcaption} 
\IEEEoverridecommandlockouts 
\usepackage[T1]{fontenc}

\usepackage{xcolor,colortbl}

\definecolor{lightlightgray}{gray}{0.9}

\title{\LARGE \bf
HEADS-UP: Head-Mounted Egocentric Dataset for Trajectory Prediction in Blind Assistance Systems
}

\author{Yasaman Haghighi$^{1}$, Celine Demonsant$^{1}$, Panagiotis Chalimourdas$^{1}$, Maryam Tavasoli Naeini$^{1}$, \\ 
Jhon Kevin Munoz$^{2}$, Bladimir Bacca$^{2}$, Silvan Suter$^{3}$, Matthieu Gani$^{3}$ and Alexandre Alahi$^{1}$%
\thanks{$^{1}$ VITA laboratory at EPFL, Lausanne, Switzerland. Email: \texttt{firstname.lastname@epfl.ch}}%
\thanks{$^{2}$ Universidad del Valle, Cali, Colombia. }%
\thanks{$^{3}$ EssentialTech Center at EPFL, Lausanne, Switzerland. Email: \texttt{firstname.lastname@epfl.ch}}%
\thanks{Corresponding author: Y. Haghighi, \texttt{yasaman.haghighi@epfl.ch}}
}
\begin{document}

\maketitle
\thispagestyle{empty}
\pagestyle{empty}

%%%%%%%%%%%%%%%%%%%%%%%%%%%%%%%%%%%%%%%%%%%%%%%%%%%%%%%%%%%%%%%%%%%%%%%%%%%%%%%%
\begin{abstract}
In this paper, we introduce HEADS-UP, the first egocentric dataset collected from head-mounted cameras, designed specifically for trajectory prediction in blind assistance systems. With the growing population of blind and visually impaired individuals, the need for intelligent assistive tools that provide real-time warnings about potential collisions with dynamic obstacles is becoming critical. These systems rely on algorithms capable of predicting the trajectories of moving objects, such as pedestrians, to issue timely hazard alerts. However, existing datasets fail to capture the necessary information from the perspective of a blind individual. To address this gap, HEADS-UP offers a novel dataset focused on trajectory prediction in this context. Leveraging this dataset, we propose a semi-local trajectory prediction approach to assess collision risks between blind individuals and pedestrians in dynamic environments. Unlike conventional methods that separately predict the trajectories of both the blind individual (ego agent) and pedestrians, our approach operates within a semi-local coordinate system—a rotated version of the camera’s coordinate system—facilitating the prediction process. We validate our method on the HEADS-UP dataset and implement the proposed solution in ROS, performing real-time tests on an NVIDIA Jetson GPU through a user study. Results from both dataset evaluations and live tests demonstrate the robustness and efficiency of our approach.
\end{abstract}

%%%%%%%%%%%%%%%%%%%%%%%%%%%%%%%%%%%%%%%%%%%%%%%%%%%%%%%%%%%%%%%%%%%%%%%%%%%%%%%%
\section{INTRODUCTION}

\begin{figure}[t]  % 'h' means place the figure approximately here in the text
    \centering
    \includegraphics[width=\linewidth]{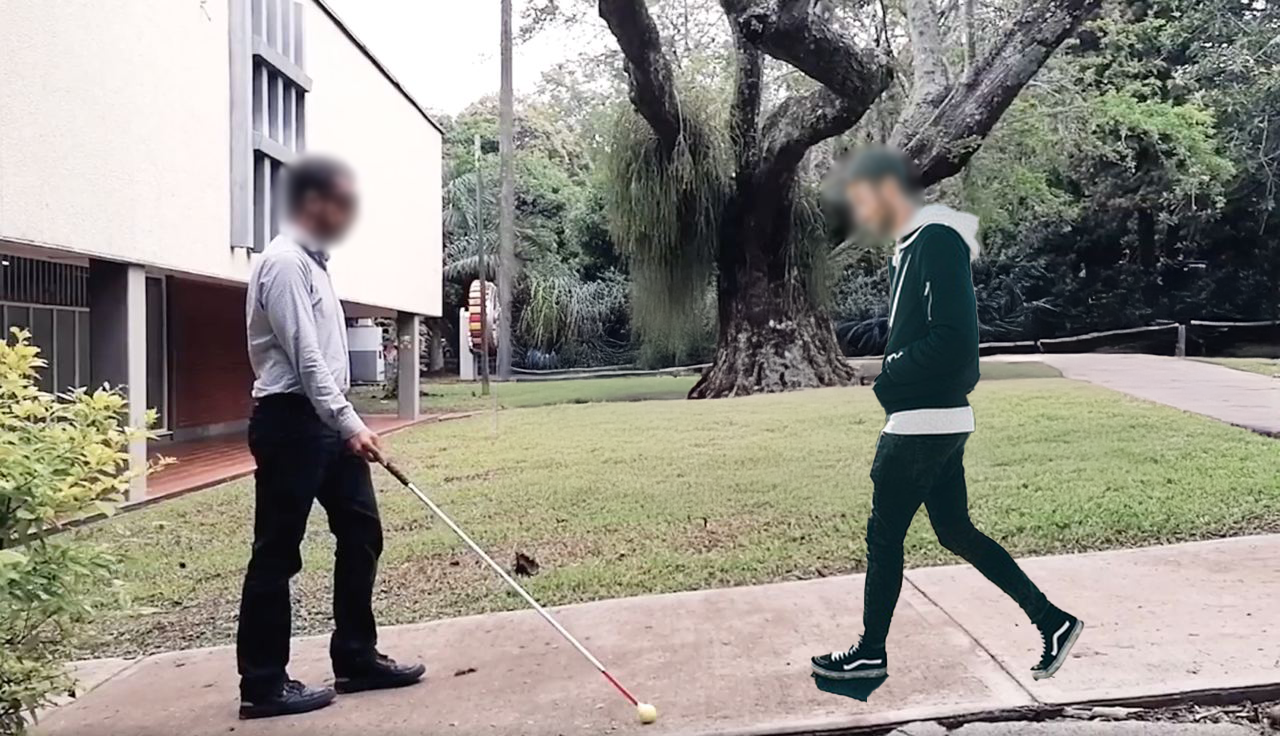}  % Set width as needed
    \caption{With the increasing number of blind and visually impaired individuals, coupled with advancements in vision-based algorithms, there is a growing need for intelligent assistive tools that can inform the blind person in advance about potential collisions with dynamic obstacles, such as pedestrians. Unlike the traditional white cane, which is limited to detecting local collisions with static objects, these systems offer enhanced navigation safety by predicting and warning of dynamic threats.}
    \label{fig:teaser}  % Reference label for the image
\end{figure}
Approximately $4\%$ of the global population is affected by blindness or severe visual impairment, including 43.3 million who are blind and 295 million experiencing moderate to severe visual impairment \cite{vision2024global}. Currently, the primary navigation aid for blind individuals is the white cane, which, while accessible and affordable, offers only limited assistance. It is primarily useful for detecting static obstacles in close proximity but provides no information about dynamic obstacles, such as pedestrians, that may pose a collision risk as shown in \Cref{fig:teaser}. This limitation is especially problematic in complex, dynamic environments, where fast-moving obstacles can lead to dangerous situations. A proper solution is to use a combination of sensors, such as cameras for online sensing, and algorithms to accurately predict the future trajectory of pedestrians and inform the blind person about potential collisions.

While previous works have suggested mounting cameras on suitcases \cite{kayukawa2020guiding} or requiring users to continuously hold a smartphone \cite{kuribayashi2022corridor}, we argue that such designs are not user-friendly. Instead, we propose a head-mounted design that offers a more practical, hands-free solution for everyday navigation. Although this approach is more user-friendly, performing trajectory prediction from head-mounted images presents significant challenges due to the constant motion of the head. To effectively tackle this problem, an egocentric dataset is required. However, existing egocentric datasets are not optimized for the trajectory prediction task. They primarily focus on individual actions \cite{grauman2022ego4d} or social interactions between people \cite{zhang2022egobody, khirodkar2023ego}, with no emphasis on potential collisions between the camera wearer and other individuals in the scene. Moreover, they lack the necessary trajectory labels for predicting movement and collision risks.

Addressing the challenge of predicting trajectories in dynamic environments requires specialized datasets that capture interactions between blind individuals and moving pedestrians. To fill this gap, we developed \textbf{HEADS-UP}, a custom dataset collected using a head-mounted stereo camera, specifically focused on scenarios where pedestrians could potentially collide with the camera wearer. \textbf{HEADS-UP} comprises more than 43,000 frames, including RGB, depth, point cloud data, IMU measurements, and trajectory labels of pedestrians. The dataset also contains approximately 1,000 individual pedestrian tracks, providing a rich source of data for trajectory prediction and collision detection tasks.

Using the \textbf{HEADS-UP} dataset, we propose a semi-local trajectory prediction baseline for blind assistance. Traditional collision detection methods typically require predicting two separate trajectories: one for the blind individual (the ego agent) and one for surrounding pedestrians. Collisions are predicted by identifying intersections between these trajectories, which necessitates knowing the positions of both the ego agent and pedestrians in a unified global coordinate system. However, we propose a simplified approach: instead of performing two independent trajectory predictions, a single prediction in a semi-local coordinate system—a rotated version of the camera’s local coordinate system—can accurately assess collision risks. Our experiments demonstrate the effectiveness and practicality of this approach.

To further validate our method, we implemented the entire pipeline in ROS and tested it in real-time on an NVIDIA Jetson GPU, demonstrating the feasibility of our approach in dynamic, real-world settings. We aim to foster further research and development in this area by making both the \textbf{HEADS-UP} dataset and the ROS implementation publicly available to the research community. For additional details, please refer to the supplementary video.

\section{RELATED WORK}
\subsection{Assistive tool for blinds}
There are many blind individuals worldwide, highlighting the need for more advanced assistive tools beyond the traditional white cane. Recent advancements in computer vision algorithms suggest that using cameras to develop assistive tools has the potential to significantly improve navigation and safety for the visually impaired. Among the proposed solutions, \cite{kuribayashi2022corridor} suggests using a combination of a smartphone and a white cane to build a 2D occupancy map of indoor environments. However, this approach focuses on static environments, and constantly holding a phone is not efficient for users. 

\cite{wang2024visiongpt} leverages recent advancements in vision-language models for video anomaly detection. While this approach can provide context to blind users, it lacks the accuracy required for detecting and predicting pedestrian trajectories, which is critical for providing real-time hazard warnings. 

For dynamic environments, \cite{kayukawa2020guiding} suggests using two RGB-D cameras, a LiDAR sensor, and an IMU, all mounted on a suitcase. This setup detects pedestrians in a global coordinate system and predicts their future trajectories. However, the design is not user-friendly due to its bulky nature. 

In contrast, we propose using a head-mounted stereo camera, which is more comfortable to wear. Additionally, we introduce a semi-local coordinate system, allowing for a single trajectory prediction. This method is more efficient than predicting two global trajectories, significantly improving the overall efficiency of the pipeline.

\subsection{Related datasets}
Among datasets captured using head-mounted cameras, Ego4D \cite{grauman2022ego4d} is one of the largest, containing over 3,000 hours of video, primarily captured with GoPro cameras. This dataset focuses mainly on daily activities and is suitable for action recognition. Mo2Cap2 \cite{xu2019mo} is another egocentric dataset captured with a fisheye camera, but it focuses on the wearer of the camera rather than other people in the environment. You2Me \cite{ng2020you2me} captures interactions between two people, while EgoBody \cite{zhang2022egobody} provides egocentric captures of two or more people interacting in indoor scenes. EgoHuman \cite{khirodkar2023ego} offers synchronized egocentric views from four people interacting in various scenarios, including activities like playing volleyball, in both indoor and outdoor environments.

For our scenario, we require a dataset not only captured from an egocentric perspective, to resemble realistic head movements, but also where the camera wearer walks through an environment with the potential for collisions with pedestrians. None of the aforementioned datasets include such cases, nor do they provide pedestrian trajectory labels.

Among the datasets commonly used for human trajectory prediction, JRDB \cite{martin2021jrdb} is captured using a social mobile manipulator, and JTA \cite{fabbri2018learning} is synthetic. While both provide trajectory labels, neither is suitable for our scenario because they are not captured from head-mounted cameras and, thus, do not account for the natural head movements that are crucial in our context.

\subsection{Pedestrian trajectory prediction}
Kalman filter \cite{r__e__kalman_1960} is a basic approach for pedestrian motion estimation tasks. The Kalman filter is a recursive algorithm that estimates the current state of a pedestrian (e.g., position and velocity) based on noisy measurements and predicts the future state of the pedestrian's trajectory using a motion model, then updating this prediction with incoming observations to minimize uncertainty. While effective for linear and smooth motions, the Kalman filter may struggle with non-linear or highly dynamic pedestrian behaviors, requiring more complex models for improved accuracy.

To improve the accuracy of trajectory predictions, early models focused on the attractive and repulsive forces between pedestrians \cite{helbing1995social} or used Bayesian inference to model human-environment interactions \cite{best2015bayesian}. Later, data-driven methods \cite{alahi2016social, kayukawa2020guiding, giuliari2021transformer, kothari2021human, chen2023unsupervised, sun2022human} were introduced to capture interactions between pedestrians, as their future trajectories are often influenced by their surroundings and other individuals. These approaches include modeling observed neighbor interactions through hidden states or directional grids.

Among various architectures like RNNs \cite{alahi2016social}, GANs \cite{gupta2018social}, and diffusion models \cite{gu2022stochastic}, state-of-the-art methods have increasingly shifted toward using transformers \cite{saadatnejad2023social}. In addition to leveraging pedestrians' 2D positional data, these models often incorporate visual cues, such as detected bounding boxes, to enhance prediction accuracy.

While these methods are highly accurate, applying them to collision detection between pedestrians and blind individuals poses challenges. Specifically, it requires representing the positions of both the pedestrian and the blind person in a unified global coordinate system. Moreover, it necessitates two separate predictions, one for the blind person and one for the pedestrian. Afterward, the system must determine whether the predicted trajectories will collide and issue a hazard warning to the blind person.

In contrast, we propose using a semi-local coordinate system to jointly predict the future relative trajectories of both agents. We demonstrate that existing approaches can perform effectively in this semi-local coordinate system, achieving results comparable to global prediction methods.

\section{DATASET}
In this section, we describe our data capture setup and collection methodology, detailing the process of annotating the data, as well as providing key dataset statistics.

\subsection{Data collection setup}
\begin{figure}[t]  
    \centering
    \includegraphics[width=0.5\linewidth]{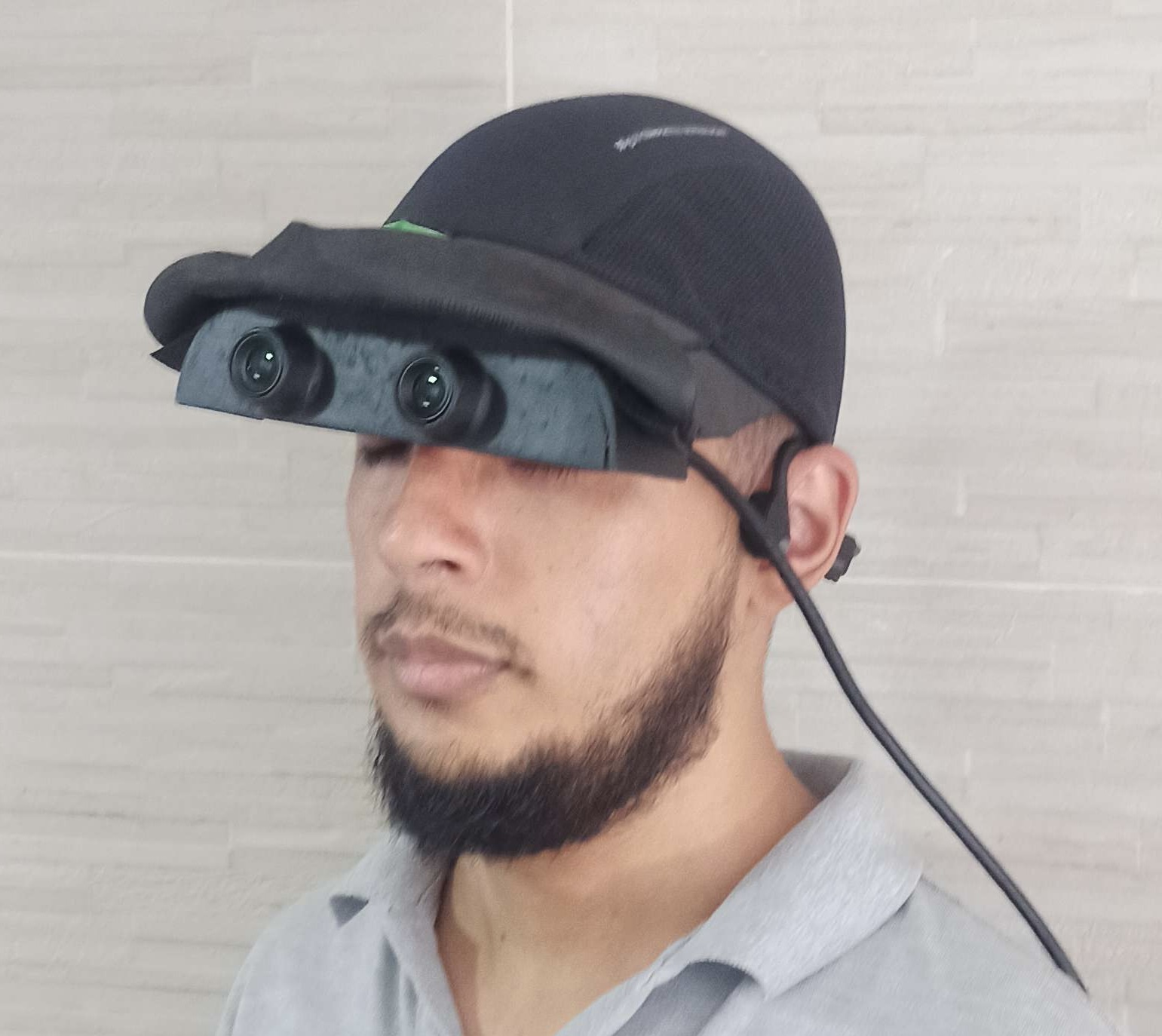}  % Set width as needed
    \caption{We use a ZED Mini stereo camera \cite{zedmini} to capture the dataset, securely mounted on a cap using a custom 3D-printed attachment to ensure stable positioning during data collection.}
    \label{fig:capture-setup}  % Reference label for the image
\end{figure}

We captured the dataset using a ZED mini camera \cite{zedmini} attached to a cap worn by the blind individual. An illustration of our capture setup is shown in \Cref{fig:capture-setup}. The ZED mini camera provided synchronized RGB images, depth maps, and IMU measurements, all recorded at a frame rate of $30$ FPS. The camera sensor specifications are summarized in \Cref{tab:camera_specs}. 

\subsection{Data collection protocol}
To tackle the task of navigation for blind individuals using a head-mounted camera, we focused on outdoor scenarios with varying levels of collision risk. The dataset captures realistic pedestrian interactions, particularly in situations where collisions with the blind individual were likely.

We considered three primary experimental setups:
\begin{itemize}
    \item \textbf{Easy setup:} The blind person walked slowly with minimal head movement, creating a controlled environment with limited distractions or challenges for trajectory prediction.
    \item \textbf{Hard setup:} The blind individual made rapid and drastic head movements, reflecting more realistic real-world behavior. This setup introduces greater complexity, as it involves more erratic head motion and pose changes, which can impact pedestrian detection and tracking.
    \item \textbf{Uncontrolled setup:} In this scenario, the blind individual navigated in an environment with multiple pedestrians, increasing the likelihood of collisions. This scenario simulates a highly dynamic and uncontrolled real-world environment with high pedestrian density and movement. 
\end{itemize}

\begin{figure*}[t!]  
    \centering
    \includegraphics[width=\linewidth]{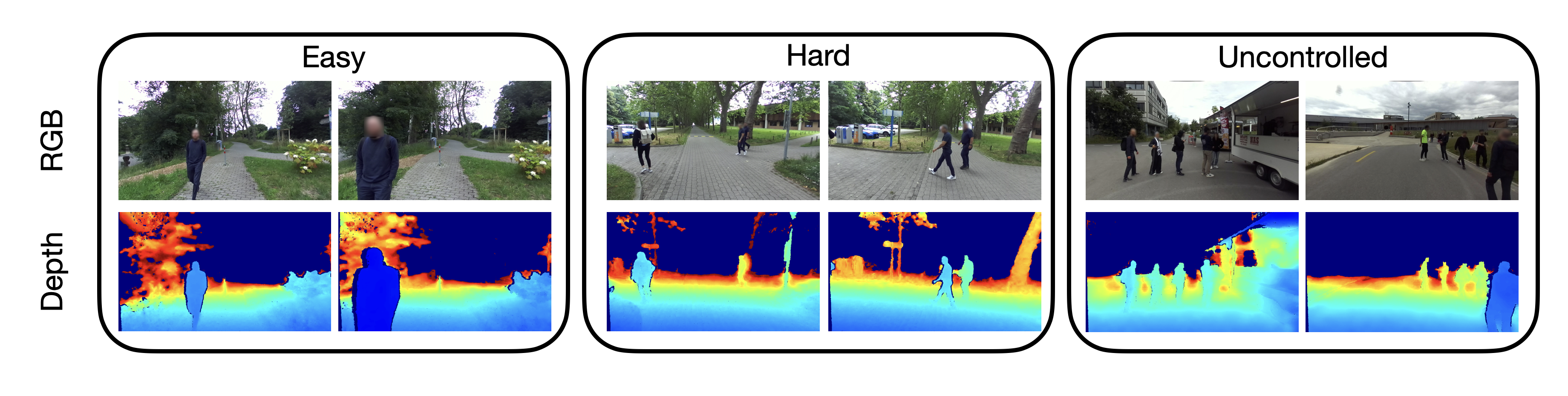}  % Set width as needed
    \caption{Example of RGB and depth images from each subset of the dataset. In the Easy subset, head movement is limited, while in the Hard subset, more drastic head movements are present. Both subsets have a controlled setup with a limited number of pedestrians. In contrast, the Uncontrolled setup features multiple pedestrians, a higher possibility of collisions, and drastic head movements. These diverse subsets enable the development and evaluation of algorithms in a variety of scenarios, facilitating research in blind navigation systems.}
    \label{fig:dataset-overview}  
\end{figure*}

Examples of RGB and depth images from each of the subsets are shown in \Cref{fig:dataset-overview}.

\subsection{Data annotation}
To obtain camera poses, we employed the Visual-Inertial Odometry (VIO) \cite{leutenegger2015keyframe} module of the ZED SDK \cite{zedsdk}. For trajectory labeling, pedestrians were first detected using the YOLOv8 \cite{JocherUltralyticsYOLO2023} object detection model. ByteTrack \cite{zhang2022bytetrack} was then employed to track the pedestrians over time, representing each by the center of their detected bounding box.

\begin{figure}[t]  
    \centering
    \includegraphics[width=0.8\linewidth]{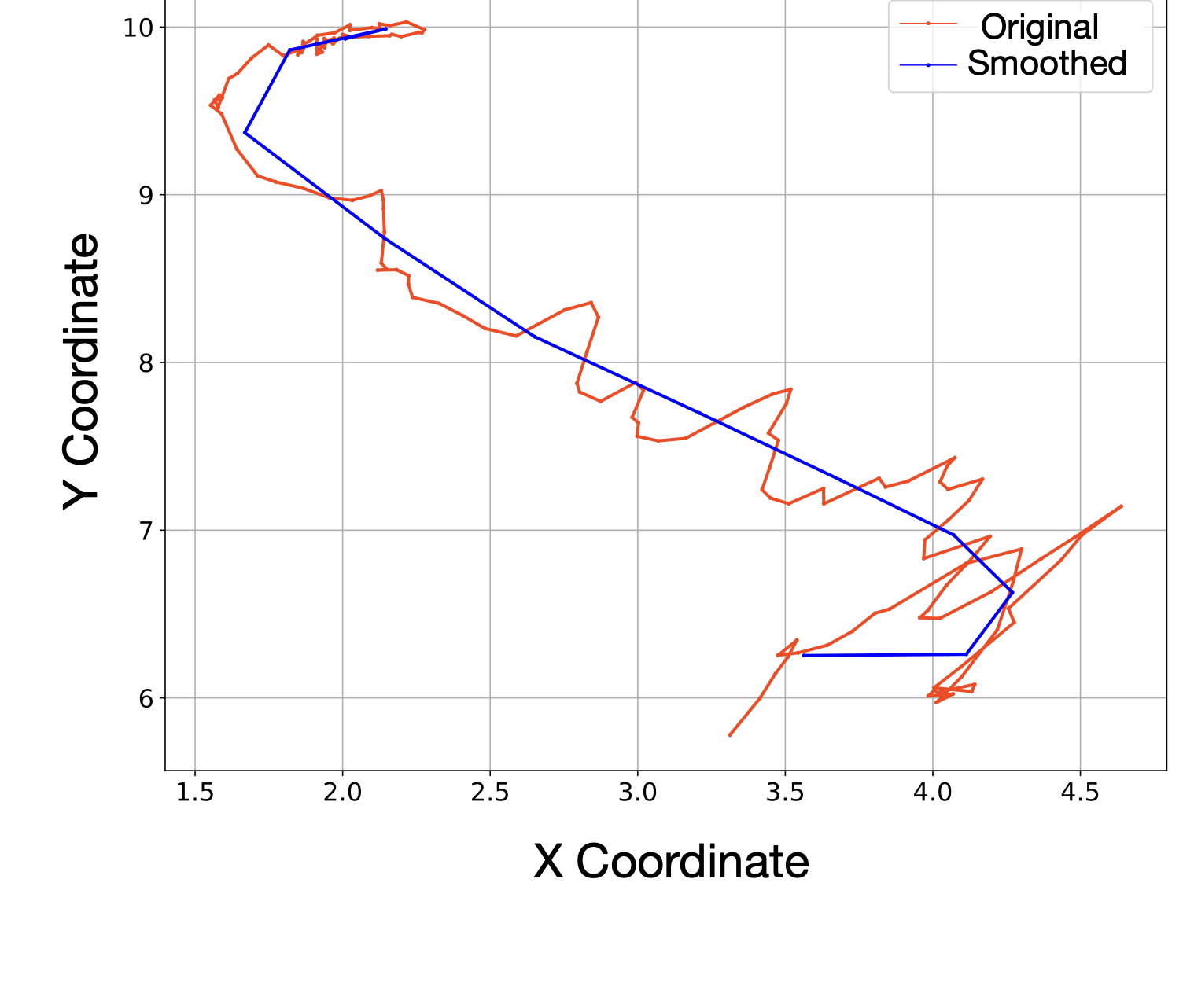}  % Set width as needed
    \caption{An example of pedestrian trajectories before (red) and after (blue) applying smoothing techniques, demonstrating reduced noise and improved trajectory reliability. The X and Y coordinates are in meters.}
    \label{fig:trajsmooth}  % Reference label for the image
\end{figure}

Due to the head-mounted nature of the system, the resulting trajectories were affected by noise from camera motion, depth capture variability, and detection model inconsistencies. To mitigate the impact of noise, we first downsampled the detected pedestrian positions to 2.5 FPS by averaging positions over consecutive frames, reducing frame-to-frame jitter. Then, we applied smoothing Kalman filter \cite{r__e__kalman_1960} to further refine the pedestrian paths. Pedestrians who were tracked for fewer than 6 frames were excluded from the dataset to ensure reliable trajectory data. An example of the trajectories before and after smoothing is shown in \Cref{fig:trajsmooth}. Finally, for privacy reasons, we blurred the faces of all pedestrians in the sequences.

\begin{table}[t]
\centering
\caption{Specifications of the ZED Mini camera used for dataset acquisition.}
\begin{tabular}{p{2cm}|p{4.2cm}}
\textbf{Specification} & \textbf{Details} \\
\hline
2x Camera & RGB, 30 Hz, $1920 \times 1080$ \\
& \\

Depth Range & 0.5 m to 25 m  \\
& \\

Depth Accuracy & $<$ 2\% up to 3 m \\
               & $<$ 4\% up to 15 m \\
               & \\

IMU & ZED Mini built-in Gyroscope and Accelerometer, 800 Hz \\

\end{tabular}
\label{tab:camera_specs}
\end{table}

\begin{table}[t]
\centering
\caption{Dataset modalities and specifications across different subsets: Easy, Hard, and Uncontrolled.}
\resizebox{\linewidth}{!}{
\Huge
\begin{tabular}{l|c|c|c|c|c|c|c}
\textbf{Subset} & \textbf{RGB} & \textbf{Depth} & \textbf{Point Cloud} & \textbf{IMU} & \textbf{Camera Pose} & \textbf{Total Frames} & \textbf{\# Agent Tracks}\\
\hline
Easy & \checkmark & \checkmark & \checkmark & \checkmark & \checkmark & 15,510 & 224\\

Hard & \checkmark & \checkmark & \checkmark & \checkmark & \checkmark & 14,038 & 169 \\

Uncontrolled & \checkmark & \checkmark & \checkmark & \checkmark & \checkmark & 13,665 & 566 \\

\end{tabular}
}
\label{tab:dataset_specs}
\end{table}

\subsection{Dataset statistics}
Our dataset comprises over 43,000 frames, including RGB images, depth data, IMU readings, point clouds, and camera poses, along with approximately 1,000 agent tracks. This provides a comprehensive and rich resource for addressing the task of navigation for blind individuals using head-mounted cameras. Further details are provided in \Cref{tab:dataset_specs}.

% \clearpage
\section{SEMI-LOCAL PREDICTION}
In this section, we present our semi-local trajectory prediction baseline. To inform the blind person about potential collisions, we need to predict the future trajectories of both the pedestrians and the blind individual. If their predicted paths intersect, we can alert the blind person to the risk of collision. Traditionally, this requires knowing the positions of both the pedestrian and the blind person in a unified global coordinate system.

However, we propose a simpler approach: using a single trajectory prediction in a semi-local coordinate system. Unlike a purely local coordinate system, where everything is relative to the camera (or the blind individual), the semi-local system takes into account the camera's rotation. This adjustment is crucial, as the blind person frequently rotates their head, and failing to consider this would result in inaccurate predictions. We run a live user study to demonstrate the effectiveness of this baseline. Below, we provide the mathematical reasoning that supports the validity of this approach:

Assume each detected and tracked pedestrian in the RGB image is represented by the center of the bounding box, denoted as \( \mathbf{p}_{uv} = \begin{bmatrix} u & v & 1 \end{bmatrix}^\top \), where \( u \) and \( v \) are the pixel coordinates of the bounding box center. Using the corresponding depth \( d_{uv} \) and the perspective camera model, we backproject the 2D pixel coordinates into camera coordinates. The backprojection from the pixel coordinates in the image plane to the camera coordinate system can be expressed as:

\begin{equation}
\mathbf{p}_{\text{camera}} = d_{uv} \cdot \mathbf{K}^{-1} \begin{bmatrix} u \\ v \\ 1 \end{bmatrix},
\end{equation}

where \( \mathbf{p}_{\text{camera}} \) is the 3D position of the pedestrian in the camera coordinate system, \( \mathbf{K} \) is the intrinsic camera matrix, and \( d_{uv} \) is the depth at pixel \( (u, v) \).

We then transform the 3D point from the camera frame to the global frame by applying the following transformation: 

\begin{equation}
\mathbf{p}_{\text{global}} = \mathbf{R} \cdot \mathbf{p}_{\text{camera}} + \mathbf{t},
\end{equation}
where \( \mathbf{R} \) is the camera rotation matrix and \( \mathbf{t} \) is the translation vector, calculated by algorithms such as \cite{leutenegger2015keyframe}. Additionally,  \( \mathbf{t} \) represents the position of the ego agent (blind individual) in the global coordinate system. 

Assume trajectory prediction algorithm is denoted by the function \( f(.) \), to detect potential collisions,  at each time step \( i \) we calculate:

\begin{equation}
f_i(\underbrace{\mathbf{R} \cdot \mathbf{p}_{\text{camera}} + \mathbf{t}}_{\text{Pedestrian}}) - f_i\underbrace{(\mathbf{t})}_{\text{Blind}} = 0.
\end{equation}

We argue that for our setup, this can be approximated as:

\begin{equation}
f_i\underbrace{(\mathbf{R} \cdot \mathbf{p}_{\text{camera}})}_{\text{Semi-Local}} \approx 0.
\end{equation}

This holds when the function \( f \) is linear, such as in the case of a Kalman filter \cite{r__e__kalman_1960}. In our experiments, we observe that this approximation extends to non-linear neural network based algorithms such as \cite{saadatnejad2023social}. 

\section{LIMITATIONS AND FUTURE WORK}
Although we have proposed a baseline for collision detection in blind navigation and demonstrated its effectiveness, there are still challenges that need to be addressed. First, semi-local trajectory prediction requires accurate estimation of the camera's rotation. In our current ROS implementation, we rely on IMU rotation measurements, which are prone to error. An end-to-end pipeline that directly estimates the possibility of collision using only RGB and depth images would be more robust and reliable. 

With the rich annotations and multimodal data in our dataset, future work could focus on training models that enhance both the efficiency and performance of collision detection for blind navigation. This would help overcome the limitations of relying on external sensors like the IMU and provide a more accurate solution. Ultimately, this could pave the way for more advanced assistive systems that offer real-time, reliable support for visually impaired individuals in navigating complex and dynamic environments.
\section{CONCLUSION}
In this paper, we introduced \textbf{HEADS-UP}, the first egocentric dataset captured using a stereo head-mounted camera, specifically designed for blind navigation. The dataset intentionally simulates scenarios with potential collisions between a blind individual and pedestrians in three Easy, Hard and Uncontrolled setups. It offers multiple modalities, including RGB, depth, IMU measurements, camera poses and trajectory labels, making it a valuable resource for the novel task of collision detection in blind navigation.

Additionally, we proposed a semi-local trajectory prediction baseline and demonstrated its effectiveness both on our dataset and in online user study evaluations. Our results show that the semi-local approach is a viable alternative to traditional global methods, providing accurate and efficient collision detection. This work paves the way for future advancements in assistive technologies for visually impaired individuals, with potential for further improvements using our dataset.

% \bibliographystyle{IEEEtran}
% \bibliography{ref}
\input{root.bbl}

\addtolength{\textheight}{-12cm}   % This command serves to balance the column lengths

%%%%%%%%%%%%%%%%%%%%%%%%%%%%%%%%%%%%%%%%%%%%%%%%%%%%%%%%%%%%%%%%%%%%%%%%%%%%%%%%

%%%%%%%%%%%%%%%%%%%%%%%%%%%%%%%%%%%%%%%%%%%%%%%%%%%%%%%%%%%%%%%%%%%%%%%%%%%%%%%%



% \section*{ACKNOWLEDGMENT}

% The preferred spelling of the word ÒacknowledgmentÓ in America is without an ÒeÓ after the ÒgÓ. Avoid the stilted expression, ÒOne of us (R. B. G.) thanks . . .Ó  Instead, try ÒR. B. G. thanksÓ. Put sponsor acknowledgments in the unnumbered footnote on the first page.

%%%%%%%%%%%%%%%%%%%%%%%%%%%%%%%%%%%%%%%%%%%%%%%%%%%%%%%%%%%%%%%%%%%%%%%%%%%%%%%%

% \begin{thebibliography}{99}

% \end{thebibliography}

\end{document}

%% file: root.bbl

%% file: root.bbl
\begin{thebibliography}{10}
\providecommand{\url}[1]{#1}
\csname url@samestyle\endcsname
\providecommand{\newblock}{\relax}
\providecommand{\bibinfo}[2]{#2}
\providecommand{\BIBentrySTDinterwordspacing}{\spaceskip=0pt\relax}
\providecommand{\BIBentryALTinterwordstretchfactor}{4}
\providecommand{\BIBentryALTinterwordspacing}{\spaceskip=\fontdimen2\font plus
\BIBentryALTinterwordstretchfactor\fontdimen3\font minus \fontdimen4\font\relax}
\providecommand{\BIBforeignlanguage}[2]{{%
\expandafter\ifx\csname l@#1\endcsname\relax
\typeout{** WARNING: IEEEtran.bst: No hyphenation pattern has been}%
\typeout{** loaded for the language `#1'. Using the pattern for}%
\typeout{** the default language instead.}%
\else
\language=\csname l@#1\endcsname
\fi
#2}}
\providecommand{\BIBdecl}{\relax}
\BIBdecl

\bibitem{vision2024global}
V.~L. E.~G. of~the Global Burden~of Disease~Study \emph{et~al.}, ``Global estimates on the number of people blind or visually impaired by cataract: a meta-analysis from 2000 to 2020,'' \emph{Eye}, vol.~38, no.~11, p. 2156, 2024.

\bibitem{kayukawa2020guiding}
S.~Kayukawa, T.~Ishihara, H.~Takagi, S.~Morishima, and C.~Asakawa, ``Guiding blind pedestrians in public spaces by understanding walking behavior of nearby pedestrians,'' \emph{Proceedings of the ACM on Interactive, Mobile, Wearable and Ubiquitous Technologies}, vol.~4, no.~3, pp. 1--22, 2020.

\bibitem{kuribayashi2022corridor}
M.~Kuribayashi, S.~Kayukawa, J.~Vongkulbhisal, C.~Asakawa, D.~Sato, H.~Takagi, and S.~Morishima, ``Corridor-walker: Mobile indoor walking assistance for blind people to avoid obstacles and recognize intersections,'' \emph{Proceedings of the ACM on Human-Computer Interaction}, vol.~6, no. MHCI, pp. 1--22, 2022.

\bibitem{grauman2022ego4d}
K.~Grauman, A.~Westbury, E.~Byrne, Z.~Chavis, A.~Furnari, R.~Girdhar, J.~Hamburger, H.~Jiang, M.~Liu, X.~Liu \emph{et~al.}, ``Ego4d: Around the world in 3,000 hours of egocentric video,'' in \emph{Proceedings of the IEEE/CVF Conference on Computer Vision and Pattern Recognition}, 2022, pp. 18\,995--19\,012.

\bibitem{zhang2022egobody}
S.~Zhang, Q.~Ma, Y.~Zhang, Z.~Qian, T.~Kwon, M.~Pollefeys, F.~Bogo, and S.~Tang, ``Egobody: Human body shape and motion of interacting people from head-mounted devices,'' in \emph{European conference on computer vision}.\hskip 1em plus 0.5em minus 0.4em\relax Springer, 2022, pp. 180--200.

\bibitem{khirodkar2023ego}
R.~Khirodkar, A.~Bansal, L.~Ma, R.~Newcombe, M.~Vo, and K.~Kitani, ``Ego-humans: An ego-centric 3d multi-human benchmark,'' in \emph{Proceedings of the IEEE/CVF International Conference on Computer Vision}, 2023, pp. 19\,807--19\,819.

\bibitem{wang2024visiongpt}
H.~Wang, J.~Qin, A.~Bastola, X.~Chen, J.~Suchanek, Z.~Gong, and A.~Razi, ``Visiongpt: Llm-assisted real-time anomaly detection for safe visual navigation,'' \emph{arXiv preprint arXiv:2403.12415}, 2024.

\bibitem{xu2019mo}
W.~Xu, A.~Chatterjee, M.~Zollhoefer, H.~Rhodin, P.~Fua, H.-P. Seidel, and C.~Theobalt, ``Mo 2 cap 2: Real-time mobile 3d motion capture with a cap-mounted fisheye camera,'' \emph{IEEE transactions on visualization and computer graphics}, vol.~25, no.~5, pp. 2093--2101, 2019.

\bibitem{ng2020you2me}
E.~Ng, D.~Xiang, H.~Joo, and K.~Grauman, ``You2me: Inferring body pose in egocentric video via first and second person interactions,'' in \emph{Proceedings of the IEEE/CVF Conference on Computer Vision and Pattern Recognition}, 2020, pp. 9890--9900.

\bibitem{martin2021jrdb}
R.~Martin-Martin, M.~Patel, H.~Rezatofighi, A.~Shenoi, J.~Gwak, E.~Frankel, A.~Sadeghian, and S.~Savarese, ``Jrdb: A dataset and benchmark of egocentric robot visual perception of humans in built environments,'' \emph{IEEE transactions on pattern analysis and machine intelligence}, vol.~45, no.~6, pp. 6748--6765, 2021.

\bibitem{fabbri2018learning}
M.~Fabbri, F.~Lanzi, S.~Calderara, A.~Palazzi, R.~Vezzani, and R.~Cucchiara, ``Learning to detect and track visible and occluded body joints in a virtual world,'' in \emph{Proceedings of the European conference on computer vision (ECCV)}, 2018, pp. 430--446.

\bibitem{r__e__kalman_1960}
R.~E. Kalman, ``A new approach to linear filtering and prediction problems,'' \emph{Journal of Basic Engineering}, vol.~82, no.~1, pp. 35--45, 1960.

\bibitem{helbing1995social}
D.~Helbing and P.~Molnar, ``Social force model for pedestrian dynamics,'' \emph{Physical review E}, vol.~51, no.~5, p. 4282, 1995.

\bibitem{best2015bayesian}
G.~Best and R.~Fitch, ``Bayesian intention inference for trajectory prediction with an unknown goal destination,'' in \emph{2015 IEEE/RSJ International Conference on Intelligent Robots and Systems (IROS)}.\hskip 1em plus 0.5em minus 0.4em\relax IEEE, 2015, pp. 5817--5823.

\bibitem{alahi2016social}
A.~Alahi, K.~Goel, V.~Ramanathan, A.~Robicquet, L.~Fei-Fei, and S.~Savarese, ``Social lstm: Human trajectory prediction in crowded spaces,'' in \emph{Proceedings of the IEEE conference on computer vision and pattern recognition}, 2016, pp. 961--971.

\bibitem{giuliari2021transformer}
F.~Giuliari, I.~Hasan, M.~Cristani, and F.~Galasso, ``Transformer networks for trajectory forecasting,'' in \emph{2020 25th international conference on pattern recognition (ICPR)}.\hskip 1em plus 0.5em minus 0.4em\relax IEEE, 2021, pp. 10\,335--10\,342.

\bibitem{kothari2021human}
P.~Kothari, S.~Kreiss, and A.~Alahi, ``Human trajectory forecasting in crowds: A deep learning perspective,'' \emph{IEEE Transactions on Intelligent Transportation Systems}, vol.~23, no.~7, pp. 7386--7400, 2021.

\bibitem{chen2023unsupervised}
G.~Chen, Z.~Chen, S.~Fan, and K.~Zhang, ``Unsupervised sampling promoting for stochastic human trajectory prediction,'' in \emph{Proceedings of the IEEE/CVF Conference on Computer Vision and Pattern Recognition}, 2023, pp. 17\,874--17\,884.

\bibitem{sun2022human}
J.~Sun, Y.~Li, L.~Chai, H.-S. Fang, Y.-L. Li, and C.~Lu, ``Human trajectory prediction with momentary observation,'' in \emph{Proceedings of the IEEE/CVF Conference on Computer Vision and Pattern Recognition}, 2022, pp. 6467--6476.

\bibitem{gupta2018social}
A.~Gupta, J.~Johnson, L.~Fei-Fei, S.~Savarese, and A.~Alahi, ``Social gan: Socially acceptable trajectories with generative adversarial networks,'' in \emph{Proceedings of the IEEE conference on computer vision and pattern recognition}, 2018, pp. 2255--2264.

\bibitem{gu2022stochastic}
T.~Gu, G.~Chen, J.~Li, C.~Lin, Y.~Rao, J.~Zhou, and J.~Lu, ``Stochastic trajectory prediction via motion indeterminacy diffusion,'' in \emph{Proceedings of the IEEE/CVF Conference on Computer Vision and Pattern Recognition}, 2022, pp. 17\,113--17\,122.

\bibitem{saadatnejad2023social}
S.~Saadatnejad, Y.~Gao, K.~Messaoud, and A.~Alahi, ``Social-transmotion: Promptable human trajectory prediction,'' \emph{arXiv preprint arXiv:2312.16168}, 2023.

\bibitem{zedmini}
\BIBentryALTinterwordspacing
Stereolabs, ``Zed mini stereo camera,'' n.d., accessed: 2024-09-15. [Online]. Available: \url{https://www.stereolabs.com/en-ch/store/products/zed-mini}
\BIBentrySTDinterwordspacing

\bibitem{leutenegger2015keyframe}
S.~Leutenegger, S.~Lynen, M.~Bosse, R.~Siegwart, and P.~Furgale, ``Keyframe-based visual--inertial odometry using nonlinear optimization,'' \emph{The International Journal of Robotics Research}, vol.~34, no.~3, pp. 314--334, 2015.

\bibitem{zedsdk}
\BIBentryALTinterwordspacing
Stereolabs, ``Zed sdk,'' n.d., accessed: 2024-09-15. [Online]. Available: \url{https://www.stereolabs.com/en-ch/developers/release}
\BIBentrySTDinterwordspacing

\bibitem{JocherUltralyticsYOLO2023}
\BIBentryALTinterwordspacing
G.~Jocher, A.~Chaurasia, and J.~Qiu, ``{Ultralytics YOLO},'' Jan 2023. [Online]. Available: \url{https://github.com/ultralytics/ultralytics}
\BIBentrySTDinterwordspacing

\bibitem{zhang2022bytetrack}
Y.~Zhang, P.~Sun, Y.~Jiang, D.~Yu, F.~Weng, Z.~Yuan, P.~Luo, W.~Liu, and X.~Wang, ``Bytetrack: Multi-object tracking by associating every detection box,'' in \emph{European conference on computer vision}.\hskip 1em plus 0.5em minus 0.4em\relax Springer, 2022, pp. 1--21.

\end{thebibliography}
